\documentclass[journal]{IEEEtran}

\newcommand{\nop}[1]{}

\usepackage{enumerate}
\usepackage{amsmath}
\usepackage{amsfonts}
\usepackage{color}
\usepackage{mathtools}
\usepackage{graphicx} 
\usepackage{subfigure}
\usepackage{epstopdf}
\usepackage{url}
\usepackage{booktabs}
\usepackage{multirow}
\usepackage{cleveref}
\usepackage{makecell}
\usepackage{balance}
\usepackage[normalem]{ulem}
\useunder{\uline}{\ul}{}

\crefname{section}{§}{§§}
\Crefname{section}{§}{§§}

\begin{document}

\title{More Behind Your Electricity Bill: a Dual-DNN Approach to Non-Intrusive Load Monitoring}

\author{Yu~Zhang,
        Guoming~Tang,
        Qianyi~Huang,
        Yi~Wang,
        Hong~Xu
\thanks{Y. Zhang, G. Tang, Q. Huang and Y. Wang are with Peng Cheng Laboratory, Shenzhen, China. H. Xu is with the Chinese University of Hong Kong, Hong Kong, China.}
\thanks{Corresponding author: G. Tang (tanggm@pcl.ac.cn).}
}

\maketitle

\begin{abstract}
Non-intrusive load monitoring (NILM) is a well-known single-channel blind source separation problem that aims to decompose the household energy consumption into itemised energy usage of individual appliances. In this way, considerable energy savings could be achieved by enhancing household's awareness of energy usage. Recent investigations have shown that deep neural networks (DNNs) based approaches are promising for the NILM task. Nevertheless, they normally ignore the inherent properties of appliance operations in the network design, potentially leading to implausible results. We are thus motivated to develop the dual Deep Neural Networks (dual-DNN), which aims to i) take advantage of DNNs' learning capability of latent features and ii) empower the DNN architecture with identification ability of universal properties. Specifically in the design of dual-DNN, we adopt one subnetwork to measure power ratings of different appliances' operation states, and the other subnetwork to identify the running states of target appliances. The final result is then obtained by multiplying these two network outputs and meanwhile considering the multi-state property of household appliances. To enforce the sparsity property in appliance's state operating, we employ median filtering and hard gating mechanisms to the subnetwork for state identification. Compared with the state-of-the-art NILM methods, our dual-DNN approach demonstrates a 21.67\% performance improvement in average on two public benchmark datasets.
\end{abstract}


\section{Introduction}
According to the statistic from UN, residential and commercial buildings consume almost 60\% of the world electricity~\cite{UNEP}. In the United States, particularly, more than 70\% of the national electricity is consumed by the building sector~\cite{US}. Meanwhile, with the explosion of high-rise building construction along with the worldwide urbanization, the building energy consumption continues increasing dramatically. Hence, energy saving in buildings is of vital importance to the reduction of overall energy consumption.

Effective and efficient energy saving in buildings can be achieved through real-time power monitoring of the end-use appliances. On the one hand, with appliances' energy consumption information in real-time, households could learn more about where their power are draining, and thus engage in sustainable energy usage campaigns more actively. It has been demonstrated that the fine-grained power consumption feedback of individual appliances could stimulate households to save 5\%$\sim$15\% energy usage~\cite{2008Feedback}. On the other hand, the disaggregated energy consumption of household appliances could be leveraged to provide references for various power management strategies~\cite{2011Disaggregated}. For example, in demand-side management programs, with real-time information of disaggregated power, utilities are able to target particular appliances (e.g., air conditioners and fridges) and suggest them to turn off or switch to energy saving modes, to shave the overall power demand in peak hours~\cite{2011Disaggregated}.

\begin{figure}[t]
\centering
\includegraphics[scale = 0.5]{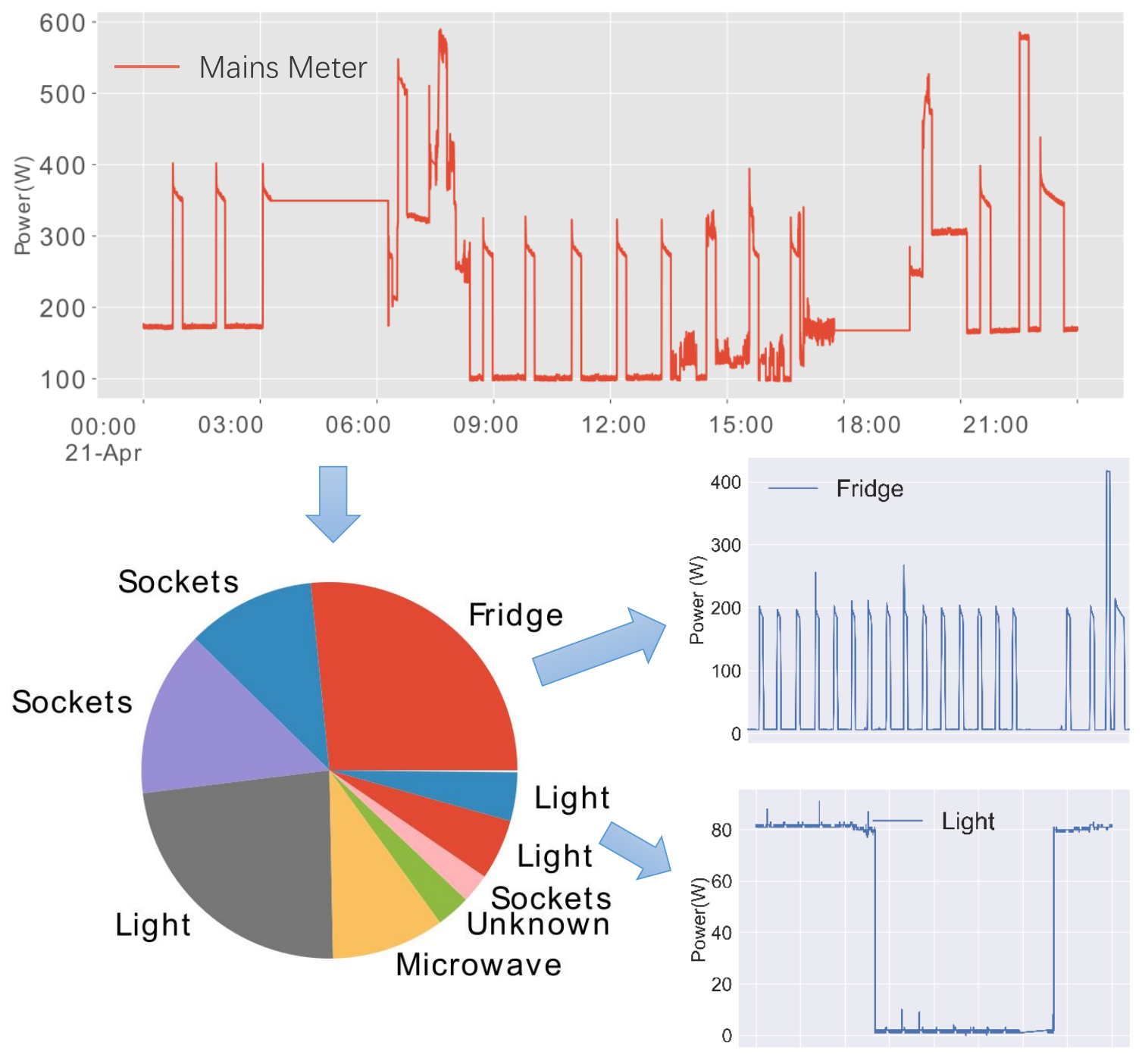}
\caption{Power readings of a whole household and two appliances based on a real-world dataset. The pie chart illustrates the energy disaggregation results of the household appliances.}
\label{fig1}
\end{figure}

Nevertheless, current power meters are incapable to reveal any fine-grained information but merely report the whole-building energy consumption. To install sensors (like smart plugs) for each sub-meter or appliance is financially prohibited, e.g., it may cost up to \$500 per house for individual sub-metering~\cite{ted}. This triggers the demand of computational techniques to infer the appliance-specific energy consumption from only the mains power reading, which is referred to as non-intrusive load monitoring (NILM)~\cite{1992Nonintrusive}. The most prominent advantage of this technique is that it can be easily adopted in existing buildings without introducing any inconvenience to households, namely being non-intrusive. However, as NILM is essentially a single-channel blind source separation (BSS) problem, i.e., to extract separated power readings of individual appliances from the single aggregated signals (as illustrated in Fig.~\ref{fig1}), it is inherently unidentifiable and theoretically intractable~\cite{zhang2016sequencetopoint}.

Recently, it has been shown that the single-channel BSS problem can be tackled by using sequence-to-sequence (seq2seq) learning with deep neural networks~
\cite{2013Deep, 2014Deep}. In particular, both deep convolutional (CNN) and recurrent neural networks (RNN) have been applied to the NILM problem~\cite{neuralnilm,zhang2016sequencetopoint}, among which the CNN architecture performs better. Specifically, the CNN structure in~\cite{zhang2016sequencetopoint} is demonstrated to be able to automatically learn instrumental features in energy disaggregation, resulting in a significant (up to 83\%) reduction rate of the estimation error. Although with decent performance, most approaches do not exploit the inherent state switching property of electric devices, and thus are not guaranteed to identify the actual operation status of end-use appliances (refer to ~\cref{sec:experiment} for detailed demonstrations). 

Based on our observations of modern appliances' power usage (see snippets of several household appliances in Fig.~\ref{fig2}), there are two common properties in their operations. We name the first one as \textbf{multi-state property}: although with some transients, the power readings of appliances are usually stable at several values, each corresponding to the power rate of one operation state. The other property from our observation is \textbf{sparsity property}: at most of the time, an appliance works under the ``stand-by'' mode, and infrequently it switches between operation states. In other words, it is impractical for most appliances to change states frequently in a short time interval. Thus their energy consumption is largely piece-wise constant over the time. However, few DNN based NILM algorithms took these important and universal properties into consideration, to say nothing of incorporating them in the network modelling. Instead, they expected that the deep neural networks could automatically learn everything (including the above properties) from scratch, which proves to be largely inefficient and sometimes impossible. Based on our analysis in~\cref{sec:observation}, DNNs are normally effective in learning the \emph{latent features} which cannot be explicitly formulated (which is also the key to DNN's success in solving general machine learning problems), whereas for the NILM problem they are insufficient in ensuring those \emph{universal features} of appliances' operations.

To address the aforementioned challenges, we borrow the idea from the multi-task neural networks and propose a dual-DNN approach to the NILM problem, by adopting one DNN for estimating power ratings of individual appliances with multiple operation modes, and the other one for identifying the correct operation states of corresponding appliances. The outputs from the dual-DNN are thus formed by multiplying the estimated power ratings with corresponding identified states. Specifically, the dual-DNN approach leverages the \emph{multi-state property} to breakdown the whole regression task into two subtasks, i.e., the power estimated task and state identification task, and guarantees the \emph{sparsity property} of appliance operation through median filtering.

As the major contribution of this work, we make the first step to incorporate universal properties in appliance operations with the design of NILM algorithms and present a novel dual-DNN approach to the NILM problem. \nop{Our proposed state identification based neural network is tailored for NILM, and thus able to extract plausible and accurate appliance usage information from aggregated mains data.} The dual-DNN approach tailored for NILM is capable to ensure unique appliance-specific properties and thus could further improve the energy disaggregation performance. We also investigate several variants of the proposed model that exploit both median filtering and hard gating mechanisms, which are more compliant with the practical settings. Compared with the state-of-the-art NILM methods, our dual-DNN approach shows 24.61\% and 27.89\% performance improvements in average on the REDD and UK-DALE datasets, respectively.

The rest of this paper is organized as follows. \cref{sec:observation} introduces how we investigate the two universal properties from modern appliances' operations. \cref{sec:preliminary} shows the basic formulation of NILM problem and sequence-to-sequence learning. In~\cref{sec:neuralnetwork}, we formally present the general framework and detailed design of the dual-DNN approach, and then present three of its variants in~\cref{sec:variants}. The implementations of this algorithm on real-world datasets and performance evaluations are shown in~\cref{sec:experiment}. \cref{sec:relatedWork} reviews the related work for NILM and \cref{sec:conclusion} concludes the paper.\nop{ and \cref{sec:acknowledgment} makes the acknowledgments.}

\section{Observation and Inspiration}\label{sec:observation}

In this section, we show our observations on two general properties from modern appliances' operations and give our insights on why and how to leverage them in solving the NILM problem with a novel dual-DNN model.

\subsection{Observations from appliance operations}

\subsubsection{Multi-State Property}
A universal property we have observed for most household appliances is their multi-state operation. Based on Fig.~\ref{fig3}, we find that typical household appliances, such as the fridge and dish washer, usually have multiple operation states, each of which corresponds to a different power rate. Normally, the multi-state appliances only work under one specific mode at any given time. Therefore, despite of some transient impulses, the power readings of an individual appliance would always be equal or approximate to the power rate of corresponding operation mode. In other words, with the power rate information of a multi-state appliance, by identifying its current state, we can readily estimate the appliance's power consumption in real-time.

\begin{figure}[t]
\centering
\includegraphics[scale = 0.48]{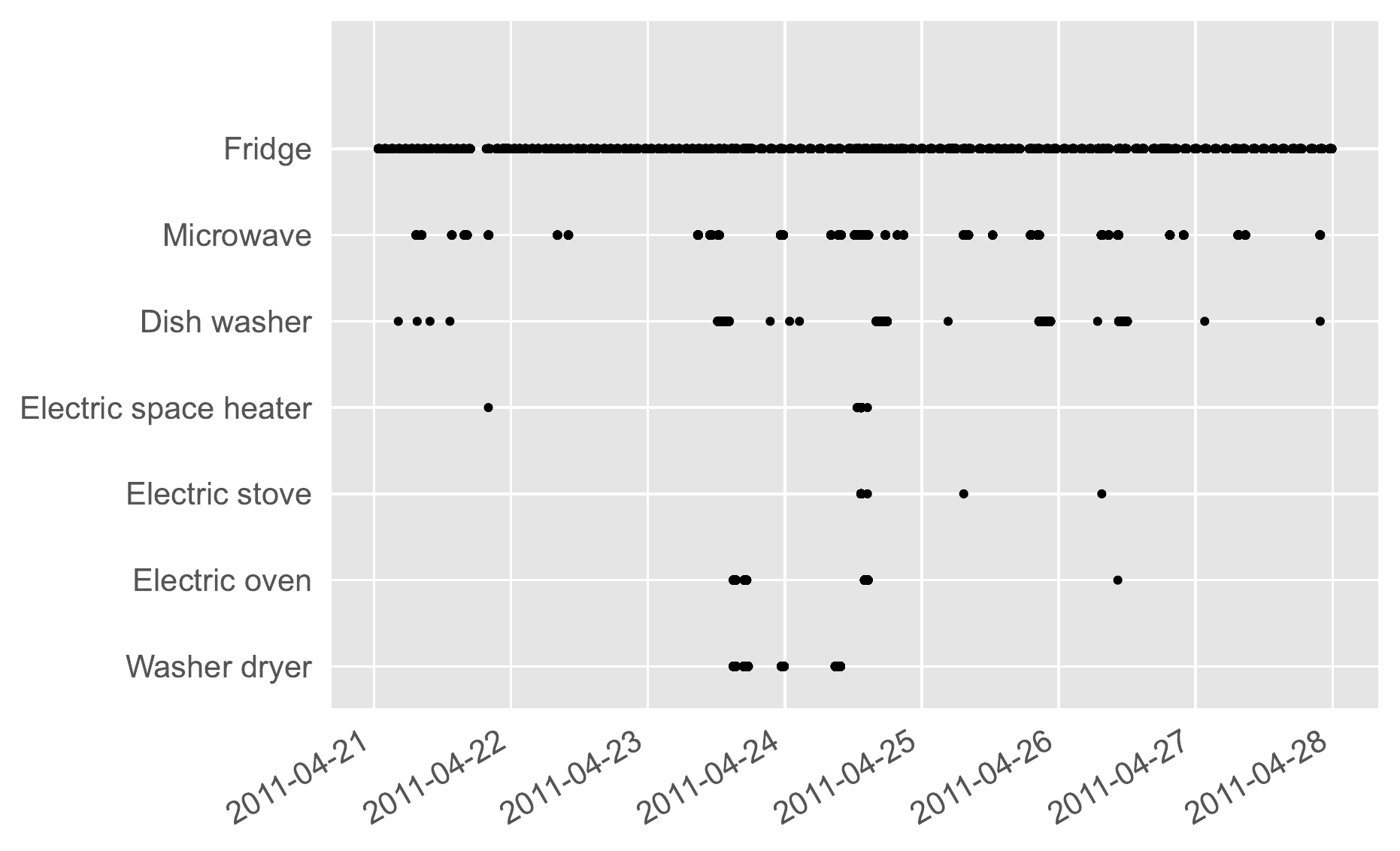}
\caption{Several household appliances' operation frequency during one week.}
\label{fig2}
\end{figure}

\subsubsection{Sparsity Property}

\begin{figure}[t]
\centering
\subfigure[Dish washer]{
\begin{minipage}[t]{0.4\linewidth}
\includegraphics[width=1\textwidth]{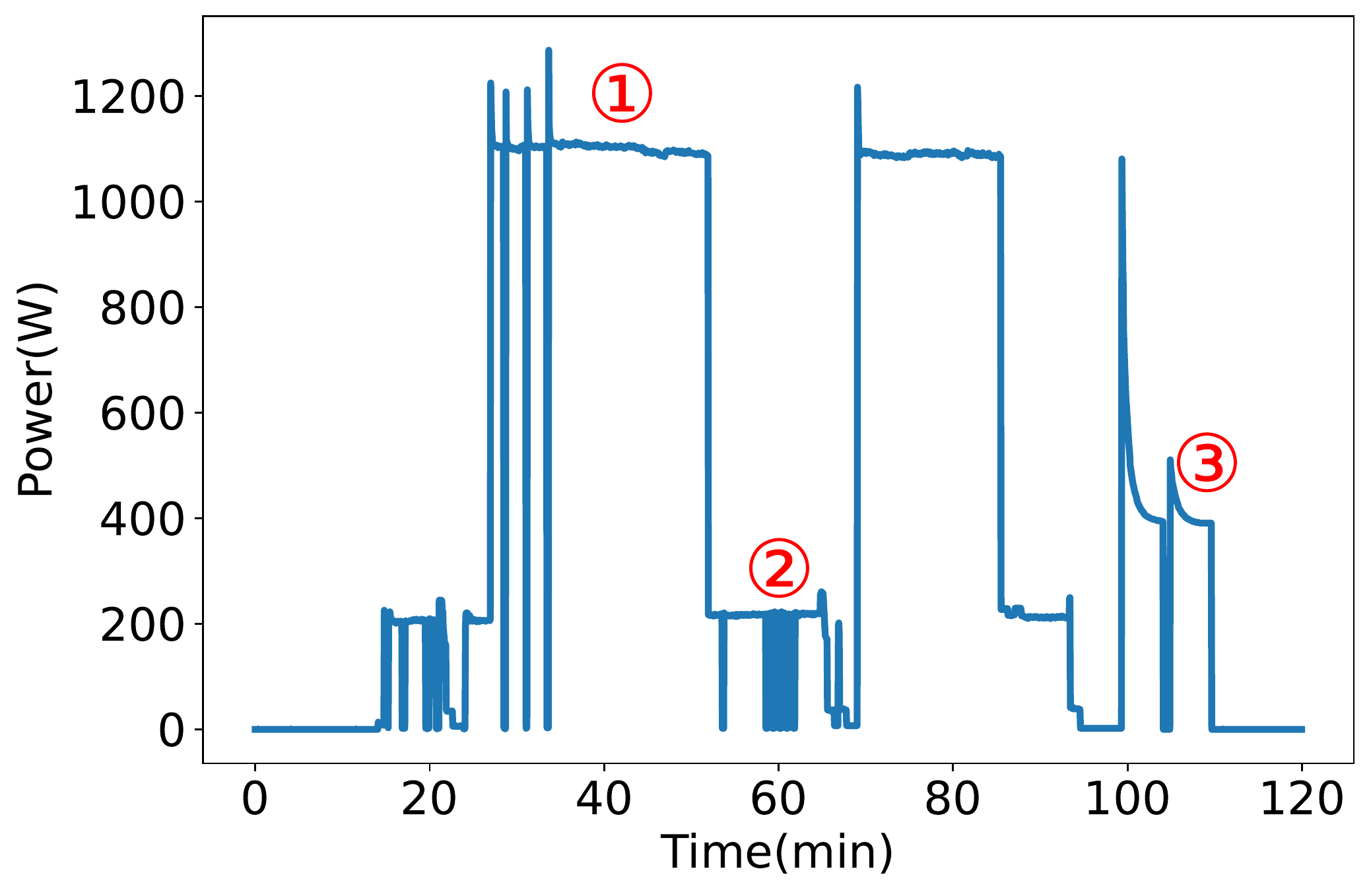}
\label{fig3a}
\end{minipage}
}%
\subfigure[Fridge]{
\begin{minipage}[t]{0.4\linewidth}
\centering
\includegraphics[width=1\textwidth]{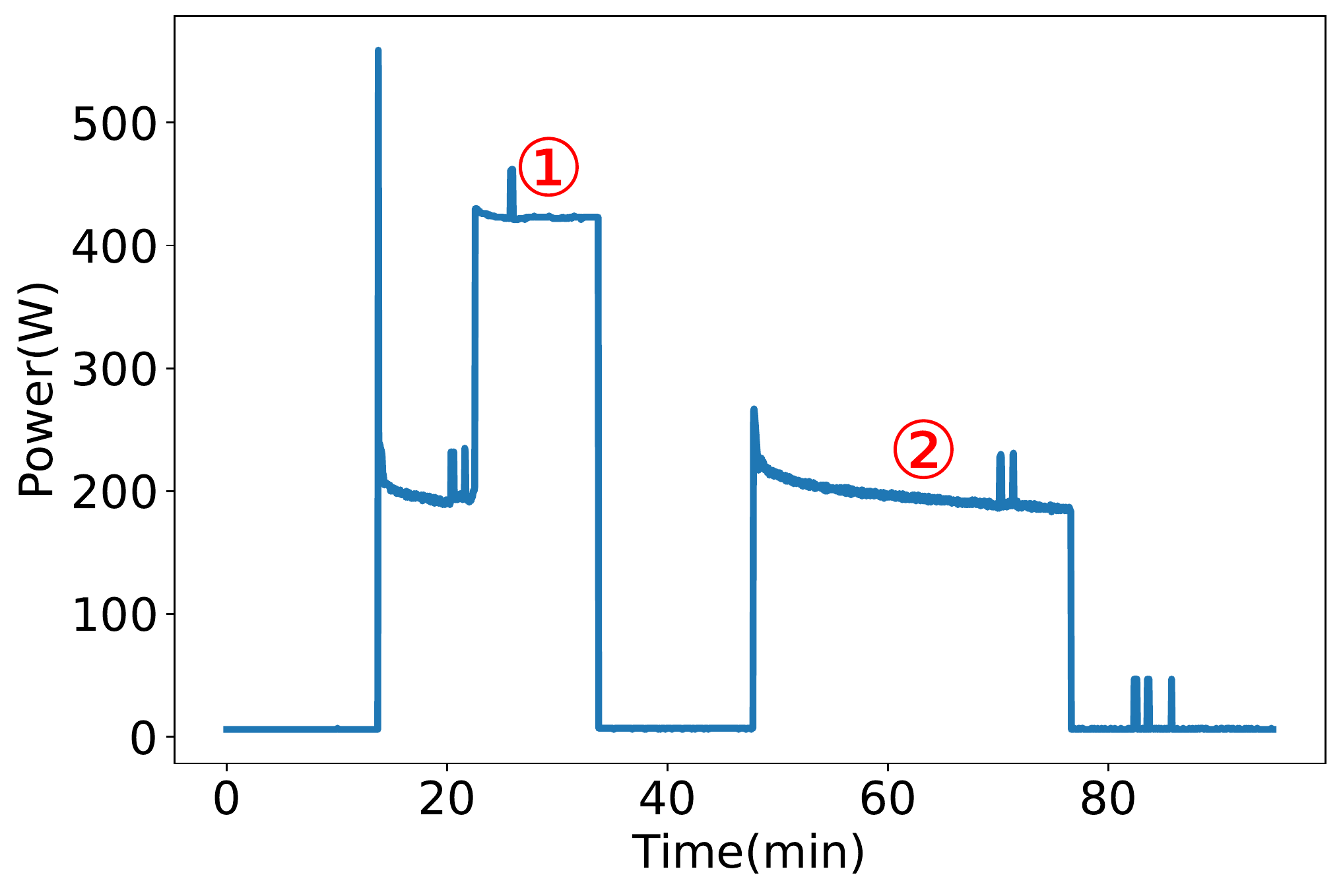}
\label{fig3b}
\end{minipage}
}%

\subfigure[Microwave]{
\begin{minipage}[t]{0.4\linewidth}
\includegraphics[width=1\textwidth]{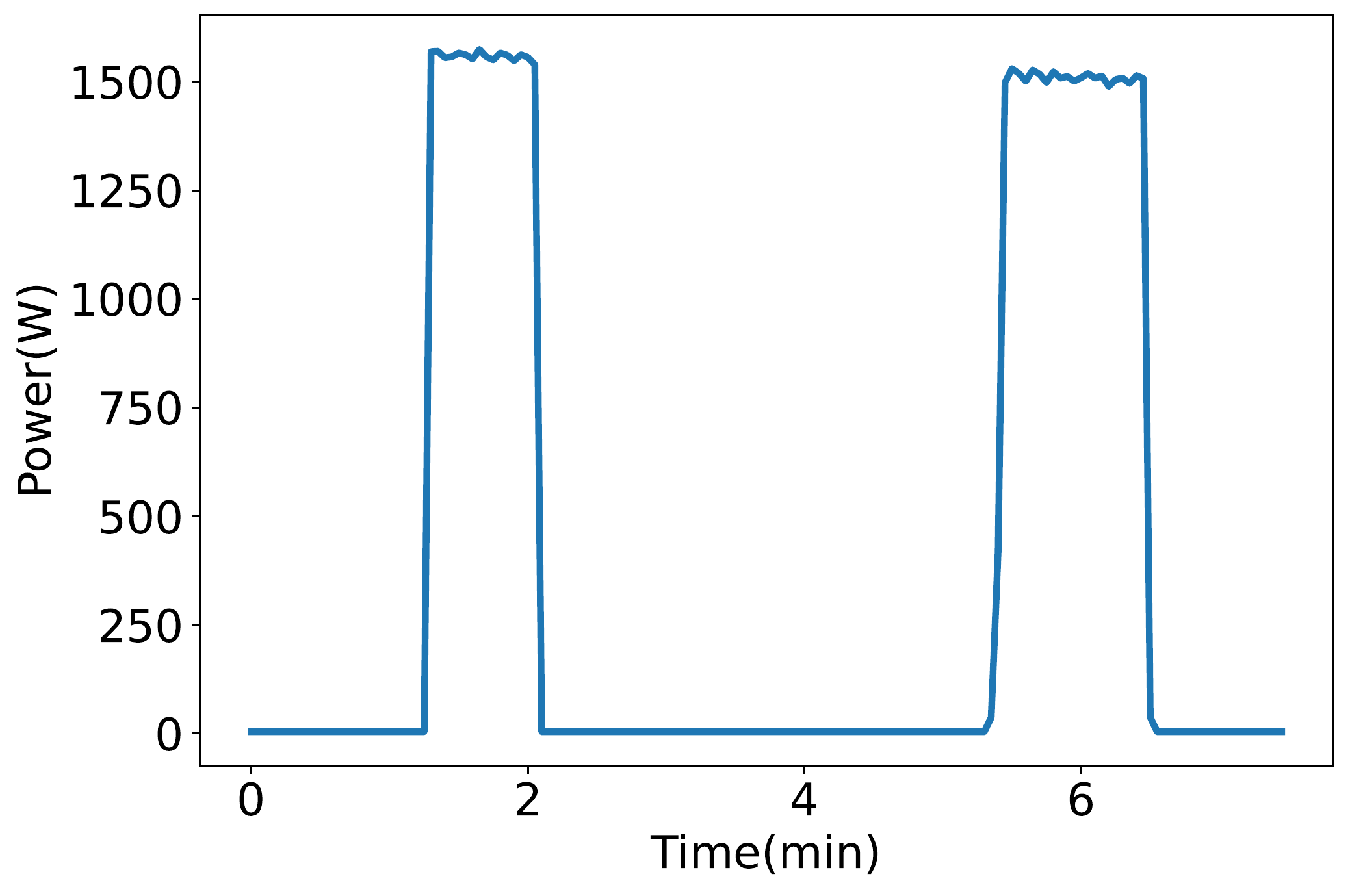}
\label{fig3c}
\end{minipage}
}%
\subfigure[Washing machine]{
\begin{minipage}[t]{0.4\linewidth}
\centering
\includegraphics[width=1\textwidth]{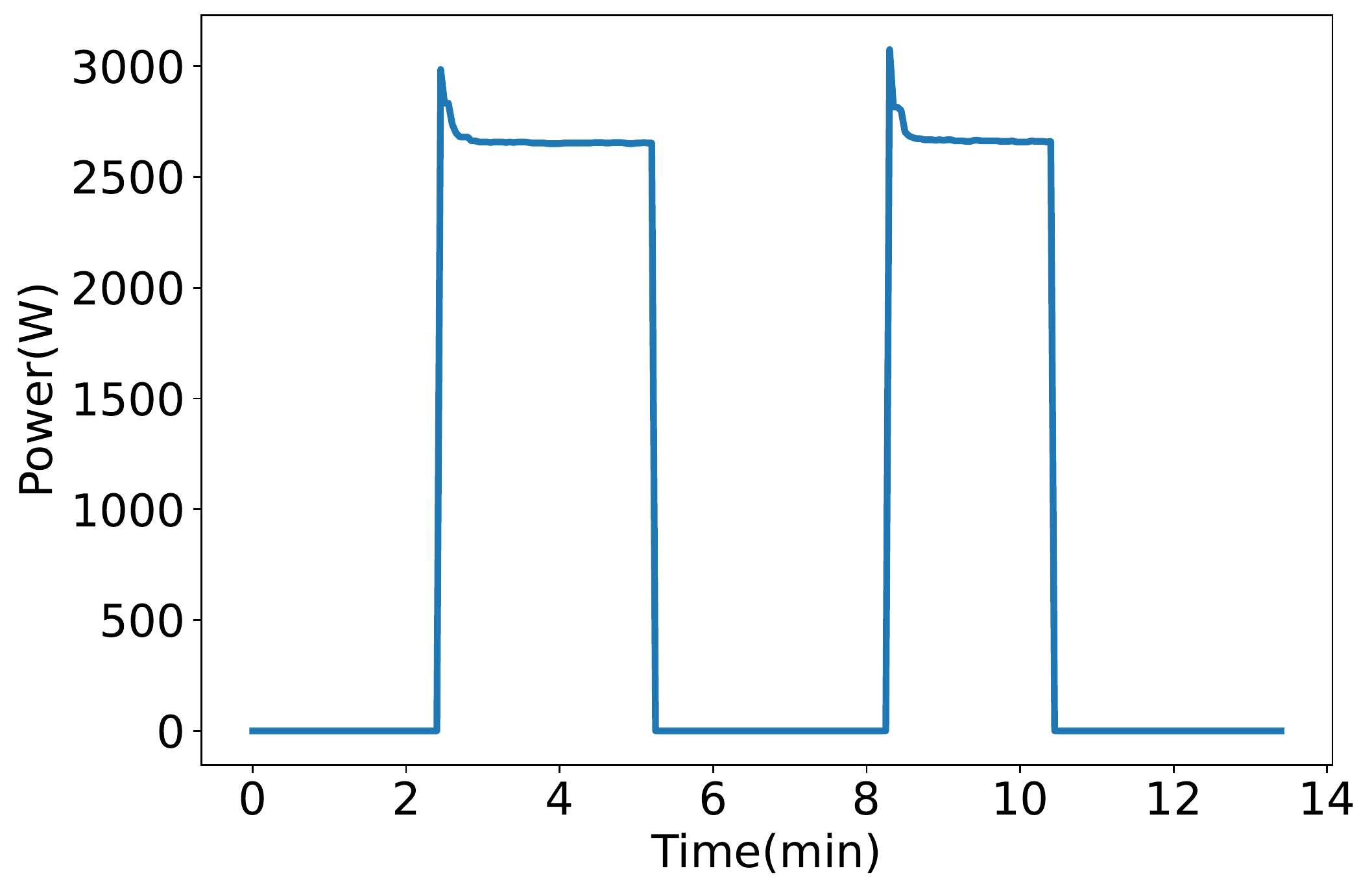}
\label{fig3d}
\end{minipage}
}%
\caption{Snippets of four typical household appliances in the REDD dataset. The different operation states of (a) the dish washer and (b) the fridge are annotated above the curves, respectively.}
\label{fig3}
\end{figure}

This property refers to the sparsity in both operation and variation of household appliances. On the one hand, appliances such as microwave and dish washer mostly operate in its OFF state (as illustrated in Fig.~\ref{fig2}). Particularly, the stand-by state could take up to 95\% of the whole appliances' working time. This implies that only a small number of appliances are under the ON state across the building, demonstrating the operation sparsity of household appliances \emph{in the spatial dimension}. On the other hand, for one specific appliance, although any of its state transitions is possible, it is rare to switch modes frequently in a short time interval. For example, as the real-world trace data shown in Fig.~\ref{fig3c} and Fig.~\ref{fig3d}, initially at the stand-by mode, the appliances first switch to their active modes and operate in these states for several sampling intervals, then return to the OFF state and stay there for a certain period of time. Thus, we can infer the operation (i.e., state switching) sparsity of appliances \emph{in the temporal dimension}. Generally speaking, a majority of the household appliances continuously stay at the stand-by mode, with each infrequently switching between its multiple states and resulting in piece-wise constant power readings over the time.

\subsection{Inspirations for dual-DNN}

In light of the multi-state property, we develop a dual-DNN framework in tackling NILM: one power estimation neural network to specifically measure power readings of different states of appliances, and the other state classification neural network to explicitly identify the current appliance's operating state. The dual-DNN framework virtually follows the multi-task DNN principles while it is tailored in this work for the specific purpose of energy disaggregation (ED)\footnote{The NILM task is also well known as energy disaggregation or ED, and we interchangeably use the two terms in this paper.}.

To leverage the sparsity property, we further enforce a median filtering mechanism into the proposed dual-DNN framework, specifically in the state classification neural network to encourage the continuity of appliance states as well as (switching) operation sparsity.

\subsection{Rationale behind dual-DNN}

As we have mentioned, the DNN was demonstrated to be able to automatically learn instrumental features for ED, including change points, typical durations and power demands of appliances~\cite{zhang2016sequencetopoint}, all of which contribute to the performance improvements in DNN based NILM algorithms. Thus, we may assume that the DNN is capable to automatically learn the aforementioned properties during model training, and thus to enforce the aforementioned properties into our model seems unnecessary. In tackling the NILM task in practice, however, things get much more complicated and the above assumption turns to be problematic.

As a matter of fact, deep neural networks are promising for the ability to extract latent features, which might be the appliance-specific features or household specialized usage patterns in NILM. Such hidden features cannot be explicitly formulated in any equations, so the only way to obtain them is through the way of deep learning. Nevertheless, household appliances also possess some general practical features, namely the aforementioned multi-state property and sparsity property. For one thing, there is no need to extract these appliance-general features through deep neural networks, as we have already learnt them as empirical knowledge in practice; for another, the performance of DNNs on learning such common features is not guaranteed, according to the results of previous DNN based NILM algorithms. 

Therefore, we develop the dual-DNN approach, which is expected to automatically learn the latent features while explicitly ensure the general features (university properties) from the appliance operations.

\section{Preliminary}\label{sec:preliminary}

\subsection{Problem formulation of NILM}

The goal of non-intrusive load monitoring is to recover the energy consumption of individual appliances from the mains readings which measure the aggregated energy consumption of the whole household. Given the aggregated power consumption for time $T$ periods as $X = (x_1, x_2, ..., x_T)$, where $x_t \in R_+$. Let $Y^i = (y^i_1, y^i_2, ..., y^i_T)$ where $y^i_t \in R_+$ denote the power readings of $i$-th appliance. Therefore, at each time $t$, $x_t$ is assumed to be the sum of several individual power readings, plus a Gaussian noise $\epsilon_t$ with zero mean and variance $\sigma^2$, which is formulated as follows:
\begin{equation}\label{eqt:1}
    x_t = \sum_i{y^i_t} + \epsilon_t
\end{equation}

Suppose that we are only interested in the top \emph{I} appliances, i.e., the ones that consume the most energy and are widely used in most of households. Then, other (unknown or low-power) appliances' energy consumption can be represented as $U = (u_1, u_2, ..., u_T)$, and Eq.~(\ref{eqt:1}) can be updated as:
\begin{equation}
    x_t = \sum_i^I{y^i_t} + u_t + \epsilon_t\label{(2)}
\end{equation}

The NILM problem is thus formulated to extract power readings of individual appliances from the mains readings, i.e., to infer $Y^1, Y^2, ..., Y^I$ from $X$.

\subsection{Seq2seq learning for NILM}

The so-called sequence-to-sequence (seq2seq) learning approach in energy disaggregation is referred to as learning a nonlinear regression between the sequence of mains readings and the sequence of a specific appliance's power readings at the same time instances~\cite{neuralnilm,zhang2016sequencetopoint,2019Subtask}. Both CNN and RNN architectures have been employed for the seq2seq learning in NILM. 

To be specific, seq2seq architectures define a neural network $f_{power}^i$ that maps a partial sequence $\tilde{x}_{t,s} = (x_{t}, ..., x_{t+s-1})$ of the mains readings (as input) to the corresponding window $\tilde{y}_{t,s}^i = (y_{t}^i, ..., y_{t+s-1}^i)$ of an individual appliance's power readings (as output). In addition, the input sequences are generally padded with two additional windows of length $w$ at the beginning and the end, respectively, to fully leverage the context information. Therefore, the input sequence is further modified as $\tilde{x}_{t,s,w} = (x_{t-w}, ..., x_{t+s+w-1})$, and the power estimation model for each individual appliance can be denoted as $f_{power}^i: \mathbb{R}_+^{s+2w} \rightarrow \mathbb{R}_+^{t}$ in the seq2seq learning.

\section{Design of Dual-DNN}\label{sec:neuralnetwork}

Recently, it has been shown in recent literature that deep neural networks are able to automatically detect specific appliance features and thus achieve better energy disaggregation performance than the optimization based NILM approaches~\cite{neuralnilm, zhang2016sequencetopoint, 2019Subtask}. However, previous DNN based approaches did not take the appliance inherent state properties into consideration and thus yielded impractical state estimations for appliances, as introduced in~\cref{sec:observation}. 

\subsection{Framework}

To better leverage the multi-state and sparsity properties of appliance operations, we propose the dual deep neural networks (dual-DNN) that are tailored to perform NILM. Specifically, we first decompose a multi-state appliance into several \emph{virtual devices} with just two states of ``ON'' and ``OFF''. Then, the sigmoid cross entropy loss is used to ensure that only one virtual device could operate at each time instant. By doing this, we are able to convert the one-at-a-time constraint to a classification problem.

\begin{figure}[t]
\centering
\includegraphics[width=0.42\textwidth]{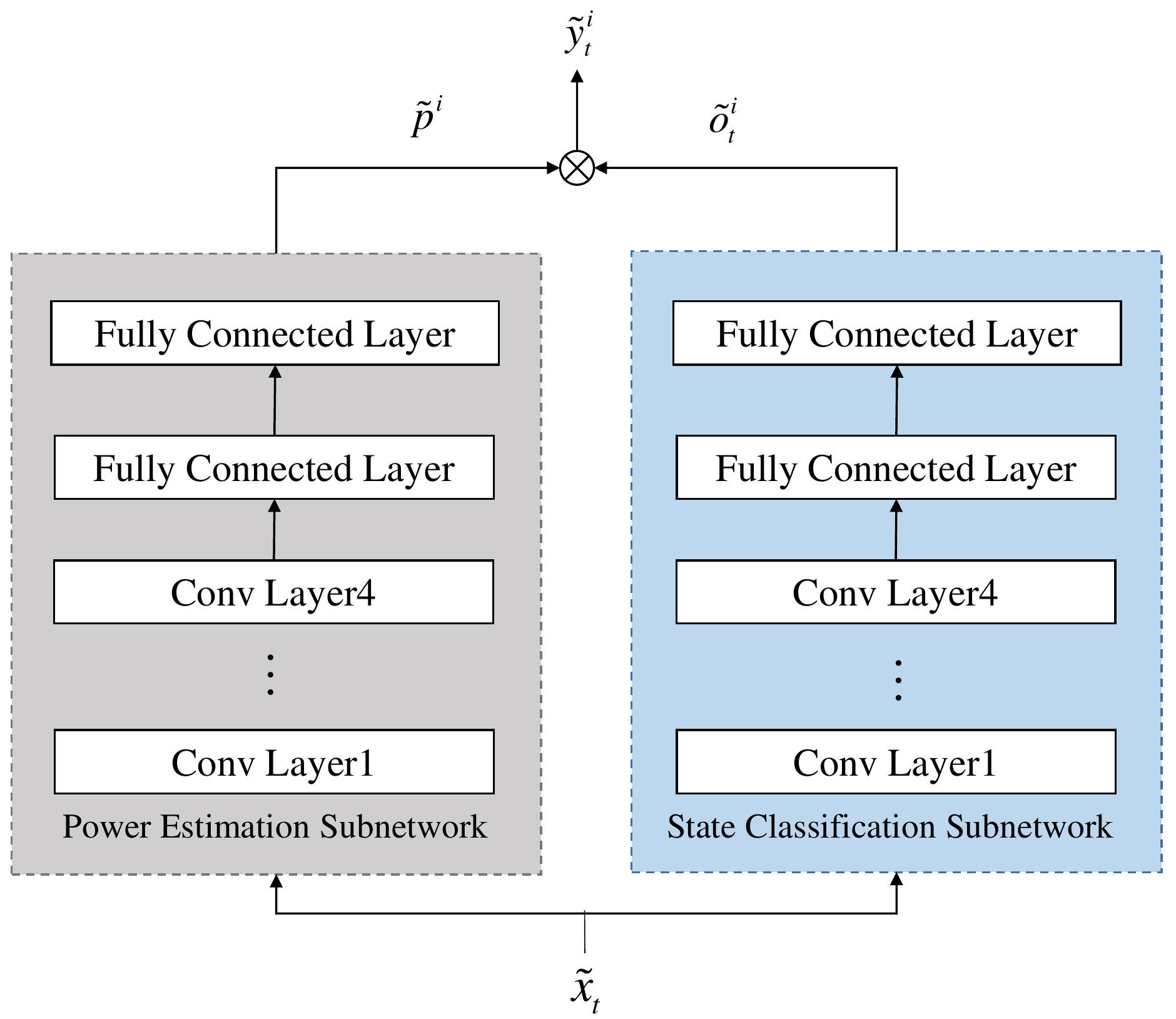}
\caption{Framework of the dual-DNN approach to NILM.}
\label{fig4}
\end{figure}

The design of the dual-DNN framework is illustrated in Fig.~\ref{fig4}. As we can see, this framework combines two DNNs: i) the power estimation subnetwork that aims to measure the power ratings of appliance in different operation states, and ii) the state classification subnetwork that is responsible for identifying the ON/OFF states of the decomposed virtual devices.

\subsection{Theoretical basis}

We then introduce the detailed design and theoretical basis of the dual-DNN model. 

\subsubsection{Left DNN Design}
For the left DNN in Fig.~\ref{fig4}, the power estimation subnetwork performs to learn a nonlinear regression between the sequence of the main power readings $\tilde{x}_t$ and the appliance power readings in different states $\tilde{p}^i$. Supposing that the $i$-th appliance has $l_i$ states (include the OFF state), the power ratings of this appliance can be represented as $\tilde{p}^i = [p_1^i, ..., p_{l_i}^i], i = 1, ..., I$. As mentioned before, we further expand the input sequence with fixed windows of length $w$ on both end sides. Therefore, for each time instant $t$, given a fixed sequence length $s$, the power estimation subnetwork uses the main power sequence $\tilde{x}_{t,s,w} = [x_{t-w}, ..., x_{t+s+w-1}]$ as the input and then estimates appliance power ratings $\tilde{p}^i = [p_1^i, ..., p_{l_i}^i]$ as the output. Overall, the appliance power estimation part can be modelled as $f_{\text{power}}^i: \mathbb{R}_+^{s+2w} \rightarrow \mathbb{R}_+^{l_i}$. Note that in spite of the input sequence length, the length of output sequence is fixed as it represents the predefined operation modes of a specific appliance. Hence, the power regression model of an individual appliance can be formulated as:
\begin{equation}\label{eqt:f-power}
    f_{\text{power}}^i(\tilde{x}_{t}) = \tilde{p}^i
\end{equation}

\subsubsection{Right DNN Design}
For the right DNN in Fig.~\ref{fig4}, the state classification subnetwork serves as the choosing unit for the main estimation task, in light of the inherent state property of household appliances. For appliance $i$, let $o_t^i(j) \in \{0,1\}, j = 1,..,l_i,$ denotes the ON/OFF state of a decomposed virtual appliance $j$ at time $t$, and:
\begin{equation}
    o_t^i(j) =
    \begin{cases} 
    1,  & \mbox{if }y_t^i\mbox{ = } p_j^i, \\
    0, & \mbox{otherwise}.
    \end{cases}
\end{equation}
Same as the power estimation subnetwork, we utilize $\tilde{x}_{t,s,w} = [x_{t-w}, \cdots, x_{t+s+w-1}]$ as the input sequence. However, unlike the fixed output length in power estimation subnetwork, the length of output sequences in state classification subnetwork largely depends on the sequence length $s$, since it aims to predict which virtual device would be active at time $t$. The virtual ON/OFF state sequence could be denoted as $\tilde{o}_{t,s,j} = (o_{t}(1),\cdots,o_{t}(l_j),\cdots,o_{t+s-1}(l),\cdots, o_{t+s-1}(l_j))$. Hence, the ON/OFF state subnetwork could be defined as $f_{\text{ON}}^i: \mathbb{R}_+^{s+2w} \rightarrow \{0,1\}^{s*l_i}$ and the mapping model is:
\begin{equation}\label{eqt:f-on}
    f_{\text{ON}}^i(\tilde{x}_{t}) = \tilde{o}_t^i
\end{equation}
Here the state identification subnetwork is indeed a classification model, with ${o}_t^i(j)$ denoting the probability that virtual appliance $j$ is at ON state at time $t$. With Eq.~(\ref{eqt:f-power}) and Eq.~(\ref{eqt:f-on}), we then can obtain the final output of the dual-DNN by:
\begin{equation}\label{eqt:f-final}
    f_{\text{output}}^i = f_{\text{power}}^i(\tilde{x}_{t}) \otimes f_{\text{ON}}^i(\tilde{x}_{t})
\end{equation}
where $f_{\text{output}}^i(t)$ is the output of dual-DNN at time $t$, and $\otimes$ represents the matrix multiplication.

\subsubsection{Loss Functions Design}
With the above design, naturally, the loss function of the dual-DNN architecture can be formulated as follows:
\begin{subequations}
\begin{align}
\mathcal{L}_{\text{output}}^i &= \frac{1}{T}\sum_{t=1}^T (y_t^i - \tilde{o}_t^i\tilde{p}^i)^2 \\
\mathcal{L}_{\text{power}}^i  &= \frac{1}{l_i}\sum_{j=1}^{l_i}(p_j^i - \tilde{p}_j^i)^2 \\
\mathcal{L}_{\text{ON}}^i &=  -\frac{1}{T}\sum_{t=1}^T\sum_{j=1}^{l_i}{o}_t^i(j)\log\tilde{o}_t^i(j)
\end{align}
\end{subequations}

Note that the power estimation subnetwork and the whole dual-DNN both aim to estimate the power consumption metric, so the mean squared error (MSE) loss is used in both loss functions $\mathcal{L}_{\text{power}}^i$ and $\mathcal{L}_{\text{output}}^i$. Meanwhile, the state identification subnetwork is responsible to learn the running state of each (virtual) appliance, namely to identify whether the appliance is currently at ON or OFF state, and therefore $\mathcal{L}_{\text{ON}}^i$ is indeed the sigmoid cross entropy loss.

For the whole dual-DNN architecture, we leverage the sum of the overall network loss and the state classification network loss for joint optimization, and thus define the whole loss function as:
\begin{equation}
    \mathcal{L} = \mathcal{L}_{\text{output}}^i + \mathcal{L}_{\text{ON}}^i
\end{equation}
Note that the cross entropy loss term $\mathcal{L}_{on}^i$ is of vital importance to the whole loss function as it not only explicitly reflects the state classification error, but also guarantee an appliance operation rule that household appliances can only operate in one mode at any given time. Hence, only through joint optimization can we obtain accurate and practical ED results.

\section{Variants of dual-DNN}\label{sec:variants}

In this section, we further modify our approach by enforcing the \emph{sparsity property} to the dual-DNN and propose several variants, which are expected to enhance the performance of original model on energy disaggregation.

\subsection{Median dual-DNN}

Based on the sparsity property of appliance operations, it is not realistic for home appliances to changes states at each time instance. Therefore, we propose to employ the median filtering, which is commonly employed in image processing to filter out pepper noise~\cite{Justusson1981Median}. Specifically, we perform the median filtering operation for the outputs from the state classification subnetwork (i.e., right DNN of the dual-DNN). Without loss of generality, we consider a particular appliance with two states: $s1$ and $s2$, and we do not expect the transition between these two states occurring frequently in short intervals, namely its power readings are expected to be piece-wise constant over the time. With $o_t$ denoting the estimated appliance state at time instance $t$, the median filtering is applied as follows:
\begin{equation}
    o_{t-L} =
    \begin{cases} 
    s_1,  & \mbox{if }o_{t-L} = s_2\mbox{ and }med(o_t, ..., o_{t-L}) = s_1 \\
    s_2,  & \mbox{if }o_{t-L} = s_1\mbox{ and }med(o_t, ..., o_{t-L}) = s_2
    \end{cases}
\end{equation}

\begin{figure*}[t]
\centering
\includegraphics[scale = 0.3]{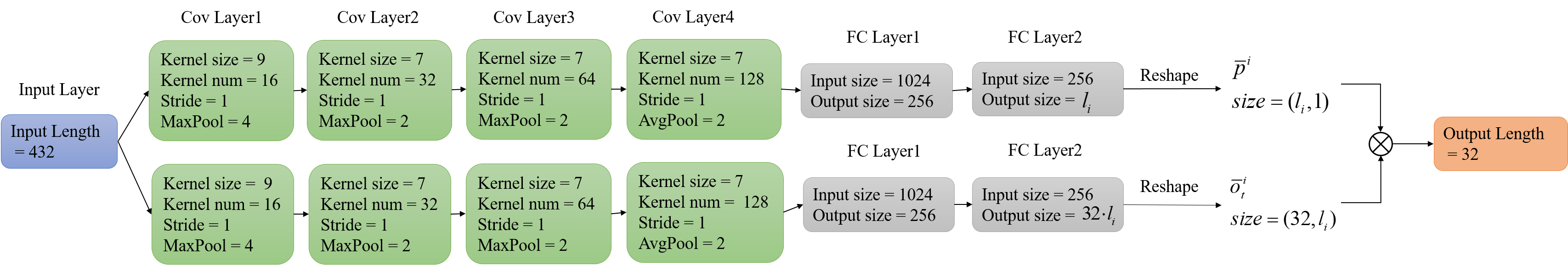}
\caption{Detailed architecture of Dual-DNN implemented in experiments for UK-DALE dataset.}
\label{fig5}
\end{figure*}

\subsection{Hard dual-DNN}

From another viewpoint, the outputs of state identification subnetwork are essentially the probabilities of virtual appliances at ON/OFF states. Therefore, instead of multiplying the estimated power ratings with the probability outputs, it is intuitive to multiply the power ratings by $1$ or $0$ (i.e., the ON or OFF state). This kind of ``hard gating'' seems more compliant to the practical appliance operation. In implementation, the hard gating goal can be achieved simply by replacing the greatest probability as $1$ and other smaller probabilities as $0$. A condition function for hard gating can be formulated accordingly:
\begin{equation}
h(x) =
\begin{cases} 
1,  & \mbox{if }x\mbox{ is the greatest probability} \\
0, & \mbox{otherwise }
\end{cases}
\end{equation}
Then the final output, given by Eq.~(\ref{eqt:f-final}), can be updated by:
\begin{equation}
    f_{\text{output}}^i = f_{\text{power}}^i(\tilde{x}_{t}) \otimes h(f_{\text{ON}}^i(\tilde{x}_{t}))
\end{equation}
Specifically, we employ the \emph{gumbel softmax}\footnote{Gumbel softmax technique is based on Gumbel-Softmax distribution that is smooth and has a well-defined gradient. In this way, the discrete one-hot-encoded categorical distributions can be further replaced by gumbel softmax samples to compute gradients~\cite{2016gumbel}.} to convert probabilities to one-hot codes, while ensuring the derivability of networks.

Furthermore, we also consider dual-DNN with both hard gating and median filtering, i.e., first modify the original outputs with hard gating function $h(x)$, and then filter out implausible impulses through median filtering. Accordingly, we name such a variant as \textbf{hard median dual-DNN}. 

\section{Experiments}\label{sec:experiment}

\begin{table}[]
\caption{Appliance parameters for the experiments. Power unit is Watt.}\label{table1}
\begin{tabular}{p{10mm}<{\centering}|p{7mm}<{\centering}|p{7mm}<{\centering}p{7mm}<{\centering}p{7mm}<{\centering}p{9mm}<{\centering}p{10mm}<{\centering}}

\hline
\multicolumn{2}{c|}{}                                                                                                         & Kettle & \begin{tabular}[c]{@{}c@{}}Micro\\ wave\end{tabular} & Fridge & \begin{tabular}[c]{@{}c@{}}Dish\\ Washer\end{tabular} & \begin{tabular}[c]{@{}c@{}}Washing\\ Machine\end{tabular} \\ \hline
\multirow{2}{*}{\begin{tabular}[c]{@{}c@{}}Window\\ Length\end{tabular}} & REDD                                               & -      & 864                                                  & 864    & 864                                                   & 864                                                       \\ \cline{2-7} 
                                                                         & \begin{tabular}[c]{@{}c@{}}UK-\\ DALE\end{tabular} & 432    & 432                                                  & 432    & 432                                                   & 432                                                       \\ \hline
\multirow{2}{*}{\begin{tabular}[c]{@{}c@{}}State\\ Number\end{tabular}}  & REDD                                               & -      & 3                                                    & 4      & 4                                                     & 3                                                         \\ \cline{2-7} 
                                                                         & \begin{tabular}[c]{@{}c@{}}UK-\\ DALE\end{tabular} & 3      & 3                                                    & 4      & 3                                                     & 4                                                         \\ \hline
\multicolumn{2}{c|}{Power Mean}                                                                                               & 700    & 500                                                  & 200    & 700                                                   & 400                                                       \\ \hline
\multicolumn{2}{c|}{Standard Deviation}                                                                                       & 1000   & 800                                                  & 400    & 1000                                                  & 700                                                       \\ \hline
\end{tabular}
\end{table}

\subsection{Datasets}

We evaluate the proposed dual-DNN for NILM tasks on two public datasets, namely REDD~\cite{2011REDD} and UK-DALE~\cite{ukdale2014} datasets, both of which contain not only the aggregate power consumption but also the individual appliance power readings.

\textbf{REDD dataset:} The REDD dataset contains power sequence data for six US houses, with 1 Hz sampling frequency for mains meter and 3 Hz for $10-25$ types of appliances meter. Since there is no kettle data, we only consider microwave, fridge, dish washer and washing machine, as these appliances are normally used in previous work ~\cite{2012fhmm, neuralnilm, zhang2016sequencetopoint, 2019Subtask}. In this way, we can compare our performance with existing solutions. 

\textbf{UK-DALE dataset:} In UK-DALE, the mains readings were recorded every 1 second and appliances power readings were recorded every 6 seconds from November 2012 to January 2015. This dataset contains aggregate power consumption and measurements of $4-54$ appliances from five UK houses. In this paper we also focus on kettle, microwave, fridge, dish washer and washing machine for the same reason mentioned above. 

\subsection{Data preprocessing}

\textbf{Filling:} After inspection, we find that it is not uncommon to see chunks of missing values, range from seconds to minutes, in mains and appliances power readings, possibly due to switched-off sensors or dead batteries. Therefore, for gaps shorter than 3 minutes, they are filled by the backward filling method, and for gaps longer than 3 minutes, they are assumed to be due to the appliance and meter being switched off and thus are filled with zeros.

\textbf{Normalization:} For both REDD and UK-DALE data, the aggregate power consumptions and individual appliances' power consumptions are preprocessed by subtracting the mean values and dividing by the standard deviations. The mean and standard deviation values of individual appliances are given in Table~\ref{table1}, both of which are obtained via statistical analysis in NILMTK~\cite{batra2019nilmtk}. After normalization, this data can be fed into DNN models for training. 

\textbf{State identification:} A rough knowledge of appliance operation modes, i.e., the number of states for each appliance, is required to build the dual-DNN. This information can be obtained from appliances' power readings in training datasets through k-means clustering or a simple visual detection. In this paper, we leverage k-means clustering to determine the state information of selected appliances in REDD and UK-DALE datasets\nop{ (see the State Number in ~\ref{table1})}. Moreover, in order to save clustering time and enhance accuracy, we first use $15$ watts as the ON state threshold, and merely cluster power rating that are larger than this threshold. The state information will be utilized in the last fully connected (FC) layers of both subnetworks to regulate the length of network outputs (refer to FC Layer2 in Fig.~\ref{fig5}). In real world, such state information can be readily acquired from user manuals of household appliances.

\renewcommand{\arraystretch}{1}
\begin{table*}[]
\caption{Experiment results on UK-DALE and REDD datasets, respectively, with best results highlighted in bold. \nop{Median DDNN is the basic dual-DNN with median filtering; Hard DDNN refers to dual-DNN with hard gating; Hard MDDNN refers to dual-DNN with both median filtering and hard gating.} }\label{table2}
\resizebox{\textwidth}{!}{
\begin{tabular}{c|c|ccccc|cccc|c}
\Xhline{2\arrayrulewidth}
\multirow{2}{*}{Metric} & \multirow{2}{*}{Model} & \multicolumn{5}{c|}{\underline{\textbf{UK-DALE}}}                                                                                                                                          & \multicolumn{4}{c|}{\underline{\textbf{REDD}}}                                                                                                                              & \multirow{2}{*}{\begin{tabular}[c]{@{}c@{}}\\AVG\\ Improve \end{tabular}} \\ 
                        &                        & Kettle        & Fridge         & Microwave      & \begin{tabular}[c]{@{}c@{}}Dish \\ Washer\end{tabular} & \begin{tabular}[c]{@{}c@{}}Washing \\ Machine\end{tabular} & Fridge         & Microwave      & \begin{tabular}[c]{@{}c@{}}Dish \\ Washer\end{tabular} & \begin{tabular}[c]{@{}c@{}}Washing \\ Machine\end{tabular} &                                                                                \\ \hline
\multirow{8}{*}{\textbf{MAE}}    & FHMM                   & 38.44         & 60.93          & 47.83          & 48.25                                                  & 66.94                                                      & 78.67          & 87.00          & 98.30                                                  & 86.24                                                      & -                                                                              \\
                        & DAE                    & 22.52         & 26.72          & 19.47          & 29.44                                                  & 18.35                                                      & 56.82          & 25.47          & 29.38                                                  & 36.25                                                      & -                                                                              \\
                        & Seq2Point              & 13.43         & 16.77          & 10.61          & 27.42                                                  & 14.55                                                      & 20.89          & 17.61          & 27.70                                                  & 22.67                                                      & -                                                                              \\
                        & SGNN                   & 11.07         & 16.71          & 9.87           & 23.16                                                  & 12.31                                                      & 22.86          & 15.98          & \textbf{14.97}                                         & 18.24                                                      & 0.00\%                                                                         \\ \cline{2-12} 
                        & Dual-DNN               & 17.60         & 17.67          & 12.15          & 17.31                                                  & 13.11                                                      & 18.47          & 13.24          & 17.20                                                  & 18.57                                                      & -1.03\%                                                                        \\
                        & Median Dual-DNN            & 9.05          & 19.72          & 11.84          & 17.04                                                  & 13.16                                                      & 16.72          & 14.18          & 18.08                                                  & 19.22                                                      & 4.24\%                                                                         \\
                        & Hard Dual-DNN              & \textbf{8.56} & \textbf{14.55} & \textbf{10.89} & 14.18                                                  & 12.46                                                      & 12.26          & \textbf{10.15} & 16.84                                                  & 18.66                                                      & 18.34\%                                                                        \\
                        & Hard Median Dual-DNN             & 8.80          & 18.30          & 11.67          & \textbf{13.40}                                         & \textbf{11.49}                                             & \textbf{10.89} & 12.67          & 16.40                                                  & \textbf{17.49}                                             & 16.57\%                                                                        \\ \hline
\multirow{8}{*}{\textbf{SAE}}    & FHMM                   & 1.85          & 0.98           & 1.04           & 2.50                                                   & 5.50                                                       & 1.46           & 1.35           & 0.98                                                   & 4.50                                                       & -                                                                              \\
                        & DAE                    & 1.35          & 0.77           & 1.14           & 1.98                                                   & 3.83                                                       & 1.06           & 1.04           & 0.78                                                   & 2.84                                                       & -                                                                              \\
                        & Seq2Point              & 1.21          & 0.56           & 0.69           & 1.59                                                   & 2.45                                                       & 0.89           & 0.86           & 0.65                                                   & 1.35                                                       & -                                                                              \\
                        & SGNN                   & 0.99          & 0.52           & 0.70           & 0.78                                                   & 2.28                                                       & 0.62           & 0.70           & 0.45                                                   & 0.83                                                       & 0.00\%                                                                         \\ \cline{2-12} 
                        & Dual-DNN               & 0.86          & 0.50           & 0.78           & 0.98                                                   & 2.66                                                       & 0.73           & 0.78           & 0.72                                                   & 0.85                                                       & -12.64\%                                                                       \\
                        & Median Dual-DNN            & 0.94          & 0.47           & 0.77           & 0.92                                                   & 2.05                                                       & 0.54           & 0.77           & 0.57                                                   & 0.83                                                       & 0.76\%                                                                         \\
                        & Hard Dual-DNN              & \textbf{0.77} & 0.37           & 0.65           & 0.83                                                   & 1.45                                                       & 0.45           & 0.62           & 0.35                                                   & \textbf{0.66}                                              & 21.82\%                                                                        \\
                        & Hard Median Dual-DNN             & 0.78          & \textbf{0.36}  & \textbf{0.59}  & \textbf{0.79}                                          & \textbf{1.28}                                              & \textbf{0.39}  & \textbf{0.57}  & \textbf{0.33}                                          & 0.67                                                       & 26.77\%                                                                        \\ 
\Xhline{2\arrayrulewidth}
\end{tabular}
}
\end{table*}

\subsection{Networks training}

As benchmarks, the performance of Factorial Hidden Markov Models (FHMM)~\cite{fhmm,2012fhmm}, denoising autoencoder (DAE)~\cite{neuralnilm} and Seq2Point~\cite{zhang2016sequencetopoint} (a variant of Seq2Seq) are evaluated. We implement the above benchmarks with NILMTK~\cite{batra2019nilmtk}, an open toolkit for analysis on non-intrusive load monitoring. As the most relevant work to ours, SGNN is also implemented according to the architectures and training details in~\cite{2019Subtask}. 

The detailed dual-DNN structure is shown in Fig.~\ref{fig5}, which adopts convolutional neural network (CNN) as basic architecture for each subnetwork, as empirical studies have demonstrated that CNNs outperform RNNs in NILM~\cite{neuralnilm}. Our network has the following hyperparameters: the batch size is 16, the leaning rate is $1.0*10^{-3}$, and the number of epoch is 10. The dual-DNN model is trained on Tesla T4 with 16GB of RAM using Pytorch. For evaluation, we leverage the last week data of each dataset for testing and the data before last week as the training set. 

As shown in Fig.~\ref{fig5}, the input sequence length for both power estimation subnetwork and state classification subnetwork is 432 for UK-DALE dataset, which is made up of partial sequence $s = 32$ and additional window $w = 200$ on both sides. The only difference in subnetwork structure is the number of neurons in the last fully connected layer, which is $l_i$ for the power estimation subnetwork and $32*l_i$ for the state classification subnetwork. Then, we further reshape the outputs of these two subnetworks and conduct matrix multiplication to obtain the final output. The input window lenth of REDD dataset is 864, with partial sequence $s = 64$ and additional window $w = 400$. The input mains sequence for both REDD and UK-DALE datasets is of $43.2$ minutes and the output sequence is $3.2$ minutes as in the previous work~\cite{2019Subtask}.

\subsection{Evaluation metrics}

We apply the mean absolute error (MAE) and signal aggregate error (SAE) to evaluate the performance of different approaches, both of which are commonly used metrics for NILM approaches. Denoting $y_t^i$ as the ground truth and $\tilde{y_t^i}$ as the estimated power consumption for appliance $i$ at time $t$, the MAE for appliance $i$ can be defined as:
\begin{equation}
    MAE^i = \frac{1}{T}\sum_{t=1}^T\left|y_t^i-\tilde{y}_t^i\right|
\end{equation}

We utilize the normalised signal aggregate error (SAE) to evaluate the aggregate estimation error over a certain period of time. Let $r^i$ and $\tilde{r^i}$ represent the ground truth and inferred total energy consumption of appliance $i$ in the total time period. Thus, the SAE can be formulated as:
\begin{equation}
    SAE^i = \frac{\left|\tilde{r}^i-r^i\right|}{r^i}
\end{equation}

A method could be accurate enough to estimate the daily appliance energy consumption (i.e., high SAE) yet may fail to achieve per-timestep prediction (i.e., low MAE). Hence, only by jointly considering MAE and SAE can we find the most practical NILM approach.

\subsection{Experiment results}

\begin{figure}[t]
\centering
\subfigure[Fridge]{
\begin{minipage}[t]{1\linewidth}
\includegraphics[width=1\textwidth]{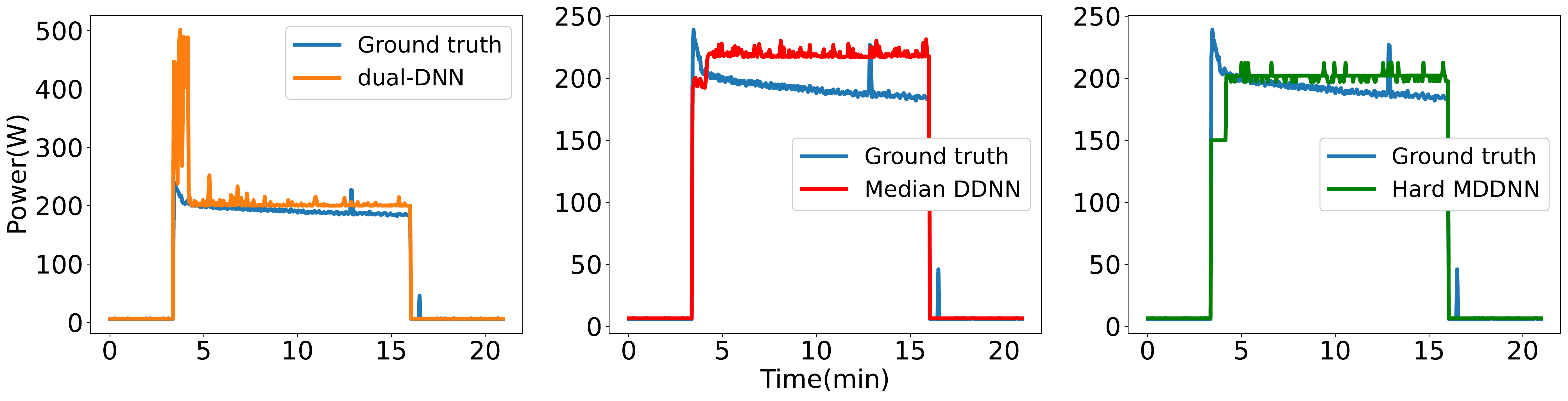}
\label{fig6a}
\end{minipage}
}%

\subfigure[Microwave]{
\begin{minipage}[t]{1\linewidth}
\includegraphics[width=1\textwidth]{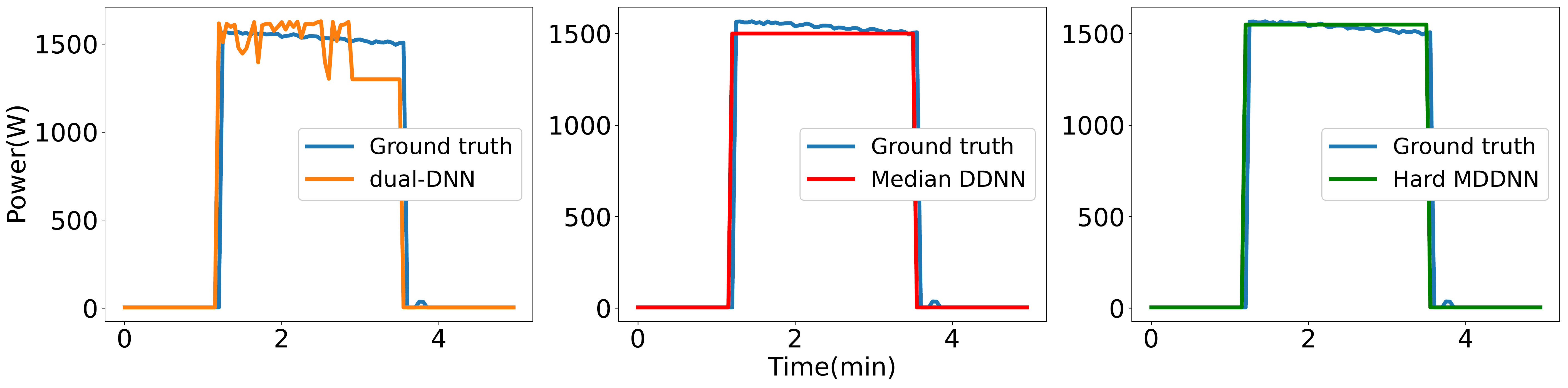}
\label{fig6b}
\end{minipage}
}%

\subfigure[Dish Washer]{
\begin{minipage}[t]{1\linewidth}
\includegraphics[width=1\textwidth]{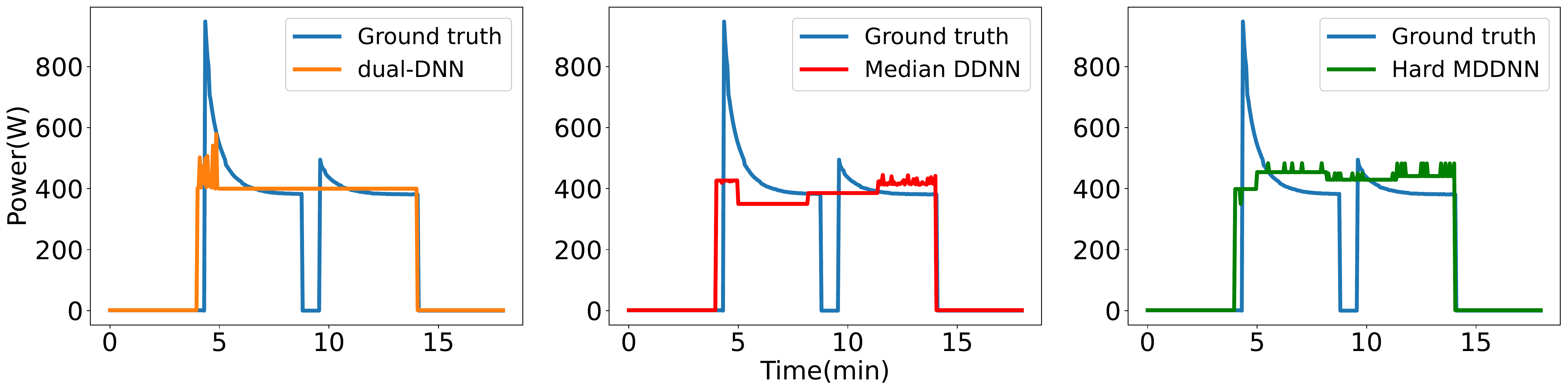}
\label{fig6c}
\end{minipage}
}%
\caption{Disaggregation results of the fridge, microwave and dish washer, respectively, in which ``MDDNN'' stands for median dual-DNN and ``HMDNN'' stands for hard median dual-DNN.}
\label{fig6}
\end{figure}

\subsubsection{Overall Results}
Table~\ref{table2} demonstrates the performance of benchmarks and our dual-DNN approach on REDD and UK-DALE datasets, respectively. The bold numbers denotes the best ED algorithms, which show that dual-DNN has surpassed the state-of-the-art performance in most cases. Specifically, our hard median dual-DNN reduces MAE by up to 16.57\% and SAE by up to 26.77\%, with improvements for 8 out of 9 cases in MAE and all 9 cases in SAE, respectively. On average, hard dual-DNN and hard median dual-DNN demonstrate approximately 16\%-27\% reduction in errors compared with the best of previous works. Median dual-DNN tends to perform worse than hard dual-DNN but still slightly outperforms the state-of-the-arts, and significant error deduction could be achieved by hard median dual-DNN. Therefore, in reality, we can choose from hard dual-DNN or hard median dual-DNN based on the appliance types.

\subsubsection{Detailed Results} 
Figure.~\ref{fig6} shows three examples of the proposed dual-DNN, median dual-DNN (MDDNN in the figure) and hard median dual-DNN (HMDNN in the figure). In the case of fridge, median filtering mainly works for filtering out the noises at the beginning of operation and hard gating mechanism helps to regulate the power estimation, i.e., to avoid the influence of higher power states. In the case of microwave, median filtering again takes the responsibility to make sure power signals piece-wise constant over time. In the case of dish washer, all three algorithms demonstrate satisfied performance in estimating power consumption of dish washer during this activation. Based on our observation, we find that the state identification subnetwork in dual-DNN is capable to learn features that indicate appliances' states, and thus succeeds in estimating typical operation durations of appliances. Meanwhile, the power estimation subnetwork does its job of estimating the power rate information of a multi-state appliance.

\subsubsection{Deep Dive} 
Seemingly, a pure dual-DNN (without median filtering and hard gating) performs worse than SGNN on average. Thus, we extend our experiments to further investigate the effectiveness of dual-DNN framework. Specifically, we implement a SGNN variant by introducing both median filtering and hard gating mechanism, and perform comparison experiments among the SGNN, the SGNN variant and the dual-DNN variant. For fast validation purpose, we only use one-house load data from UK-DALE and REDD, respectively. The results from new experiments are summarized in Table.~\ref{table3}, from which we find that: i) the SGNN variant outperforms the original SGNN with a $\sim$18\% performance improvement (12.50\% for MAE and 22.67\% for SAE, respectively), and ii) our approach (dual-DNN with median filtering and hard gating) outperforms the SGNN variant by $\sim$11\% (8.32\% for MAE and 13.12\% for SAE, respectively). Therefore, we can conclude that the strength of our approach indeed comes from two factors: the dual-DNN block, and the median filtering and hard gating block. The former is designed to extract appliances’ \textbf{latent} features through deep neural networks, and the latter part is designed to enforce the \textbf{general} features in appliances’ operations. The experimental results also verified that, only through the combination of the two parts can we achieve plausible and accurate energy disaggregation results. 

\renewcommand{\arraystretch}{1}
\begin{table*}[]
\caption{Results from UK-DALE and REDD datasets, respectively, with best results highlighted in bold. \nop{Median DDNN is the basic dual-DNN with median filtering; Hard DDNN refers to dual-DNN with hard gating; Hard MDDNN refers to dual-DNN with both median filtering and hard gating.}}
\label{table3}
\resizebox{\textwidth}{!}{
\begin{tabular}{c|c|ccccc|cccc|c}
\Xhline{2\arrayrulewidth}
\hline
\multirow{2}{*}{Metric} & \multirow{2}{*}{Model} & \multicolumn{5}{c|}{{\ul UK-DALE} (One-house Data)}                                                                                                              & \multicolumn{4}{c|}{{\ul REDD} (One-house Data)}                                                                                                         & \multirow{2}{*}{\begin{tabular}[c]{@{}c@{}}AVG\\ Improve\end{tabular}} \\
                        &                        & Kettle & Fridge & Microwave & \begin{tabular}[c]{@{}c@{}}Dish\\ Washer\end{tabular} & \begin{tabular}[c]{@{}c@{}}Washing\\ Machine\end{tabular} & Fridge & Microwave & \begin{tabular}[c]{@{}c@{}}Dish\\ Washer\end{tabular} & \begin{tabular}[c]{@{}c@{}}Washing\\ Machine\end{tabular} &                                                                        \\ \hline
\multirow{3}{*}{\textbf{MAE}}    & SGNN                   & 15.24  & 20.77  & 12.74     & 26.81                                                 & 16.19                                                     & 25.78  & 19.05     & 17.38                                                 & 22.15                                                     & -                                                                      \\
                        & Hard Median SGNN       & 13.07  & 17.51  & 10.65     & 23.51                                                 & 15.31                                                     & 22.74  & 16.38     & 15.14                                                 & 19.78                                                     & 8.32\%                                                                \\
                        & Hard Median Dual-DNN   & \textbf{11.56}  & \textbf{15.72}  & \textbf{10.44}     & \textbf{19.44}                                                 & \textbf{14.73}                                                    & \textbf{19.86}  & \textbf{14.78}     & \textbf{16.03}                                                & \textbf{18.71}                                                    & 12.50\%                                                                 \\ \hline
\multirow{3}{*}{\textbf{SAE}}    & SGNN                   & 1.37   & 0.86   & 1.01      & 1.28                                                  & 2.75                                                      & 0.98   & 1.10      & 0.79                                                  & 1.18                                                      & -                                                                      \\
                        & Hard Median SGNN       & 0.95   & 0.77   & 0.88      & 0.90                                                  & 2.47                                                      & 0.60   & 0.88      & 0.57                                                  & 0.73                                                      & 13.12\%                                                                \\
                        & Hard Median Dual-DNN   & \textbf{0.81}   & \textbf{0.60}   & \textbf{0.76}      & \textbf{0.78}                                                  & \textbf{2.18}                                                      & \textbf{0.51}   & \textbf{0.77}      & \textbf{0.49}                                                  & \textbf{0.70}                                                      & 22.67\%                                                                \\ \hline
\Xhline{2\arrayrulewidth}
\end{tabular}}
\end{table*}

\section{Related Work}\label{sec:relatedWork}

Non-intrusive load monitoring or energy disaggregation, was first introduced by George Hart in 1992~\cite{1992Nonintrusive}. Since that, various approaches have been proposed to solve this single channel BSS problem. The NILM methods can be broadly classified as i) optimization based approaches and ii) DNN based approaches.

\textbf{Optimization based Approaches:} To begin with, optimization based algorithms generally define the load aggregation task as an optimization problem, and usually employ techniques such as evolutionary algorithms~\cite{2013EvoNILM, egarter2015load}, linear and nonlinear integer programming approaches~\cite{2008IP, 2016median}. The prior knowledge of appliances, including the operation states and corresponding power ratings which can be easily acquired from users' manual, is frequently leveraged to obtain the optimal solution. The ED performance of optimization-based algorithms largely depends on the objective functions and related constraints. Apart from the most commonly employed objective function - least square error (LSE) between measured and approximated aggregate power consumption~\cite{2013EvoNILM}, there are several enhancements via median filtering~\cite{2016median}, convex penalty term~\cite{piga2015sparse} and linear-programming based refinement~\cite{liu2019linear}. However, such optimization based algorithms fail in practical settings as they assume the aggregate energy consumption equals the sum of considered appliances' usage. But the thing is we cannot input all the household appliance information into the model and literature shows that the energy consumption from unknown sources, such as living room usage and electric cars, can take up 51.86\% of total energy~\cite{2019tree}, which limit the practicability and effectiveness of optimization based ED algorithms. 

\textbf{DNN based Approaches:} Deep learning approaches are demonstrated to be promising for NILM with its excellent performance. In~\cite{neuralnilm}, the authors take the first step to leverage various deep learning models including convolutional neural networks (CNN), recurrent neural networks (RNN), and Denoising Autoencoder (DAE) in NILM problems. In light of sequence-to-sequence (seq2seq) learning, the authors in ~\cite{zhang2016sequencetopoint} suggest a sequence-to-point learning based on CNN structure to map the single mid-point of appliance's power reading with mains reading. Also inspired by seq2seq learning, the authors in~\cite{2019VRNN} propose a novel deep generative architecture for performing sequence-to-many-sequence learning, i.e., mains power consumption to several appliances' power consumption. Then, in~\cite{2019Subtask}, the researchers begun to utilize ON/OFF state property of electric devices to serve as a ``gating unit'', namely refinement, for their regression results. While demonstrating significant performance improvement, none of these deep learning based NILM methods leverage inherent operation properties of end-use appliances, leading to rather implausible energy breakdown results.

In summary, the optimization based approaches utilize appliance state information to obtain accurate ED results but suffer from practical settings; the DNN based approaches leverage deep learning to enhance NILM performance yet fail to take appliances' properties into account. To the best of our knowledge, our solution is the first DNN based approach that aims to measure the appliance's operation state and leverages underlying appliance properties to enhance ED performance.

\balance

\section{Conclusions}\label{sec:conclusion}

In this paper, we investigated the well-known non-intrusive load monitoring problem. Our data analysis revealed that the operations of household appliances generally have notable multi-state and sparsity properties, which can be exploited for NILM. We were inspired to develop a dual-DNN for energy disaggregation. Specifically, we introduced a multi-task neural network framework that consists of two parallel subnetworks, one aims to estimate the power ratings of appliances at different states, and the other is responsible to identify current operation states of devices. Empirical evaluations on benchmark datasets and algorithms validated the effectiveness and practicability of our solution. The dual-DNN approach presented in this work could be potentially applied to other learning tasks that can be divided into one estimation problem and the other classification problem.

\ifCLASSOPTIONcaptionsoff
  \newpage
\fi

\bibliographystyle{IEEEtran}  
\bibliography{reference}  

\begin{thebibliography}{10}
\providecommand{\url}[1]{#1}
\csname url@samestyle\endcsname
\providecommand{\newblock}{\relax}
\providecommand{\bibinfo}[2]{#2}
\providecommand{\BIBentrySTDinterwordspacing}{\spaceskip=0pt\relax}
\providecommand{\BIBentryALTinterwordstretchfactor}{4}
\providecommand{\BIBentryALTinterwordspacing}{\spaceskip=\fontdimen2\font plus
\BIBentryALTinterwordstretchfactor\fontdimen3\font minus
  \fontdimen4\font\relax}
\providecommand{\BIBforeignlanguage}[2]{{%
\expandafter\ifx\csname l@#1\endcsname\relax
\typeout{** WARNING: IEEEtran.bst: No hyphenation pattern has been}%
\typeout{** loaded for the language `#1'. Using the pattern for}%
\typeout{** the default language instead.}%
\else
\language=\csname l@#1\endcsname
\fi
#2}}
\providecommand{\BIBdecl}{\relax}
\BIBdecl

\bibitem{UNEP}
UNEP, ``Energy efficiency for buildings,''
  \url{https://www.euenergycentre.org/images/unep info sheet - ee
  buildings.pdf}, 2016.

\bibitem{US}
A.~T.~S. ENERGY, ``Overview,'' \url{https://www.ase.org/initiatives/buildings},
  2018.

\bibitem{2008Feedback}
C.~Fischer, ``Feedback on household electricity consumption: a tool for saving
  energy?'' \emph{Energy Efficiency}, vol.~1, no.~1, pp. 79--104, 2008.

\bibitem{2011Disaggregated}
J.~Froehlich, E.~Larson, S.~Gupta, G.~Cohn, M.~Reynolds, and S.~Patel,
  ``Disaggregated end-use energy sensing for the smart grid,'' \emph{IEEE
  pervasive computing}, vol.~10, no.~1, pp. 28--39, 2010.

\bibitem{ted}
TED, ``The energy detective,'' \url{http://bit.ly/28UKP62}, 2020.

\bibitem{1992Nonintrusive}
G.~W. Hart, ``Nonintrusive appliance load monitoring,'' \emph{Proceedings of
  the IEEE}, vol.~80, no.~12, pp. 1870--1891, 1992.

\bibitem{zhang2016sequencetopoint}
C.~Zhang, M.~Zhong, Z.~Wang, N.~Goddard, and C.~Sutton, ``Sequence-to-point
  learning with neural networks for nonintrusive load monitoring,'' 2016.

\bibitem{2013Deep}
E.~M. Grais, M.~U. Sen, and H.~Erdogan, ``Deep neural networks for single
  channel source separation,'' in \emph{2014 IEEE International Conference on
  Acoustics, Speech and Signal Processing (ICASSP)}.\hskip 1em plus 0.5em minus
  0.4em\relax IEEE, 2014, pp. 3734--3738.

\bibitem{2014Deep}
P.-S. Huang, M.~Kim, M.~Hasegawa-Johnson, and P.~Smaragdis, ``Deep learning for
  monaural speech separation,'' in \emph{2014 IEEE International Conference on
  Acoustics, Speech and Signal Processing (ICASSP)}.\hskip 1em plus 0.5em minus
  0.4em\relax IEEE, 2014, pp. 1562--1566.

\bibitem{neuralnilm}
\BIBentryALTinterwordspacing
J.~Kelly and W.~Knottenbelt, ``Neural nilm: Deep neural networks applied to
  energy disaggregation,'' in \emph{Proceedings of the 2nd ACM International
  Conference on Embedded Systems for Energy-Efficient Built Environments}, ser.
  BuildSys '15.\hskip 1em plus 0.5em minus 0.4em\relax New York, NY, USA:
  Association for Computing Machinery, 2015, p. 55–64. [Online]. Available:
  \url{https://doi.org/10.1145/2821650.2821672}
\BIBentrySTDinterwordspacing

\bibitem{2019Subtask}
C.~Shin, S.~Joo, J.~Yim, H.~Lee, and W.~Rhee, ``Subtask gated networks for
  non-intrusive load monitoring,'' \emph{Proceedings of the AAAI Conference on
  Artificial Intelligence}, vol.~33, pp. 1150--1157, 2019.

\bibitem{Justusson1981Median}
B.~Justusson, ``Median filtering: Statistical properties,'' in
  \emph{Two-Dimensional Digital Signal Prcessing II}.\hskip 1em plus 0.5em
  minus 0.4em\relax Springer, 1981, pp. 161--196.

\bibitem{2016gumbel}
E.~Jang, S.~Gu, and B.~Poole, ``Categorical reparameterization with
  gumbel-softmax,'' \emph{arXiv preprint arXiv:1611.01144}, 2016.

\bibitem{2011REDD}
J.~Z. Kolter and M.~J. Johnson, ``Redd: A public data set for energy
  disaggregation research,'' in \emph{Workshop on data mining applications in
  sustainability (SIGKDD), San Diego, CA}, vol.~25, no. Citeseer, 2011, pp.
  59--62.

\bibitem{ukdale2014}
J.~Kelly and W.~Knottenbelt, ``The uk-dale dataset, domestic appliance-level
  electricity demand and whole-house demand from five uk homes,''
  \emph{Scientific data}, vol.~2, no.~1, pp. 1--14, 2015.

\bibitem{2012fhmm}
J.~Z. Kolter and T.~Jaakkola, ``Approximate inference in additive factorial
  hmms with application to energy disaggregation,'' in \emph{Artificial
  intelligence and statistics}.\hskip 1em plus 0.5em minus 0.4em\relax PMLR,
  2012, pp. 1472--1482.

\bibitem{batra2019nilmtk}
N.~Batra, R.~Kukunuri, A.~Pandey, R.~Malakar, R.~Kumar, O.~Krystalakos,
  M.~Zhong, P.~Meira, and O.~Parson, ``Towards reproducible state-of-the-art
  energy disaggregation,'' in \emph{Proceedings of the 6th ACM International
  Conference on Systems for Energy-Efficient Buildings, Cities, and
  Transportation}, 2019, pp. 193--202.

\bibitem{fhmm}
Z.~Ghahramani and M.~I. Jordan, ``Factorial hidden markov models,''
  \emph{Machine learning}, vol.~29, no.~2, pp. 245--273, 1997.

\bibitem{2013EvoNILM}
D.~Egarter and W.~Elmenreich, ``Evonilm: Evolutionary appliance detection for
  miscellaneous household appliances,'' in \emph{Proceedings of the 15th annual
  conference companion on Genetic and evolutionary computation}, 2013, pp.
  1537--1544.

\bibitem{egarter2015load}
------, ``Load disaggregation with metaheuristic optimization,'' in
  \emph{Energieinformatik}, 2015, pp. 1--12.

\bibitem{2008IP}
K.~Suzuki, S.~Inagaki, T.~Suzuki, H.~Nakamura, and K.~Ito, ``Nonintrusive
  appliance load monitoring based on integer programming,'' in \emph{2008 SICE
  Annual Conference}.\hskip 1em plus 0.5em minus 0.4em\relax IEEE, 2008, pp.
  2742--2747.

\bibitem{2016median}
M.~Z.~A. Bhotto, S.~Makonin, and I.~V. Bajic, ``Load disaggregation based on
  aided linear integer programming,'' \emph{IEEE Transactions on Circuits \&
  Systems II Express Briefs}, vol.~64, no.~7, pp. 792--796, 2016.

\bibitem{piga2015sparse}
D.~Piga, A.~Cominola, M.~Giuliani, A.~Castelletti, and A.~E. Rizzoli, ``Sparse
  optimization for automated energy end use disaggregation,'' \emph{IEEE
  Transactions on Control Systems Technology}, vol.~24, no.~3, pp. 1044--1051,
  2015.

\bibitem{liu2019linear}
Y.~Liu, Y.~Sun, and B.~Li, ``A modified ip-based nilm approach using appliance
  characteristics extracted by 2-sax,'' \emph{IEEE Access}, vol.~7, pp.
  48\,119--48\,128, 2019.

\bibitem{2019tree}
Y.~Jia, N.~Batra, H.~Wang, and K.~Whitehouse, ``A tree-structured neural
  network model for household energy breakdown,'' in \emph{The World Wide Web
  Conference}, 2019, pp. 2872--2878.

\bibitem{2019VRNN}
G.~Bejarano, D.~Defazio, and A.~Ramesh, ``Deep latent generative models for
  energy disaggregation,'' \emph{Proceedings of the AAAI Conference on
  Artificial Intelligence}, vol.~33, pp. 850--857, 2019.

\end{thebibliography}

\nop{\begin{IEEEbiography}{Michael Shell}
Biography text here.
\end{IEEEbiography}

\begin{IEEEbiographynophoto}{John Doe}
Biography text here.
\end{IEEEbiographynophoto}
}

\end{document}